\pdfoutput=1

\documentclass[11pt]{article}

\usepackage[preprint]{acl}

\usepackage{amsmath,amsfonts,bm}

\def\eqref#1{equation~\ref{#1}}

\def\1{\bm{1}}

\def\vw{{\bm{w}}}

\def\mF{{\bm{F}}}

\def\mH{{\bm{H}}}

\def\mY{{\bm{Y}}}
\def\mZ{{\bm{Z}}}

\DeclareMathAlphabet{\mathsfit}{\encodingdefault}{\sfdefault}{m}{sl}
\SetMathAlphabet{\mathsfit}{bold}{\encodingdefault}{\sfdefault}{bx}{n}

\DeclareMathOperator*{\argmax}{arg\,max}

\usepackage{times}
\usepackage{latexsym}

\usepackage{multirow}
\usepackage{booktabs}

\usepackage[T1]{fontenc}

\usepackage[utf8]{inputenc}

\usepackage{microtype}

\usepackage{inconsolata}

\usepackage{graphicx}

\title{\textit{EnriCo}: Enriched Representation and Globally Constrained Inference for Entity and Relation Extraction}

\author{Urchade Zaratiana$^{1,2}$, Nadi Tomeh$^2$, Yann Dauxais$^{1}$, Pierre Holat$^{1,2}$, Thierry Charnois$^2$ \\
$^1$ FI Group, 
$^2$ LIPN, CNRS UMR 7030, France \\{\tt zaratiana@lipn.fr} \\
Code: \texttt{\textbf{https://github.com/urchade/enrico}}}

\begin{document}
\maketitle
\begin{abstract}
Joint entity and relation extraction plays a pivotal role in various applications, notably in the construction of knowledge graphs. Despite recent progress, existing approaches often fall short in two key aspects: richness of representation and coherence in output structure. These models often rely on handcrafted heuristics for computing entity and relation representations, potentially leading to loss of crucial information. Furthermore, they disregard task and/or dataset-specific constraints, resulting in output structures that lack coherence. In our work, we introduce \textit{EnriCo}, which mitigates these shortcomings. Firstly, to foster rich and expressive representation, our model leverage attention mechanisms that allow both entities and relations to dynamically determine the pertinent information required for accurate extraction. Secondly, we introduce a series of decoding algorithms designed to infer the highest scoring solutions while adhering to task and dataset-specific constraints, thus promoting structured and coherent outputs. 
Our model demonstrates competitive performance compared to baselines when evaluated on Joint IE datasets.
\end{abstract}

\section{Introduction}
Joint entity and relation extraction is a pivotal task in Natural Language Processing (NLP), aiming to identify entities (such as ``\textit{Person}'' or ``\textit{Organization}'') within raw text and to discern the relationships between them (such as ``\textit{Work\_for}''). This process forms the cornerstone for numerous applications, including the construction of Knowledge Graphs \citep{7358050}. Traditionally, this task is tackled via pipeline models that independently trained and implemented entity recognition and relation extraction, often leading to error propagation \citep{10.1007/10704656_11, Nadeau2007ASO}. The advent of deep learning has facilitated the development of end-to-end and multitask models for this task, enabling the utilization of shared representations and the simultaneous optimization of loss functions for both tasks.  \citep{wadden-etal-2019-entity, wang-lu-2020-two, Zhao_Yan_Cao_Li_2021, zhong-chen-2021-frustratingly, yan-etal-2021-partition}. Despite this advancement, these models essentially remain pipeline-based, with entity and relation predictions executed by separate classification heads, thereby ignoring potential interactions between these tasks.


While end-to-end models have been proposed \citep{lin-etal-2020-joint}, they often resort to hand-coded operations like concatenation for computing entity and relation representations, thereby limiting their flexibility. Moreover, these representations ignore potential inter-span and inter-relation interactions, as well as their interactions with the input text. Integrating these interactions could enrich the representations by preserving valuable contextual information overlooked during pooling operations. Moreover, existing approaches tend to overlook the structured nature of the output. In many real-world scenarios, the relationships between entities follow certain patterns or constraints, which may vary depending on the domain or dataset. However, current models typically treat entity and relation extraction as separate classification tasks without considering these constraints explicitly. Consequently, the extracted entities and relations may lack coherence or violate domain-specific rules, limiting the utility of the extracted knowledge.

\begin{figure*}[t]
    \centering
\includegraphics[width=0.95\textwidth]{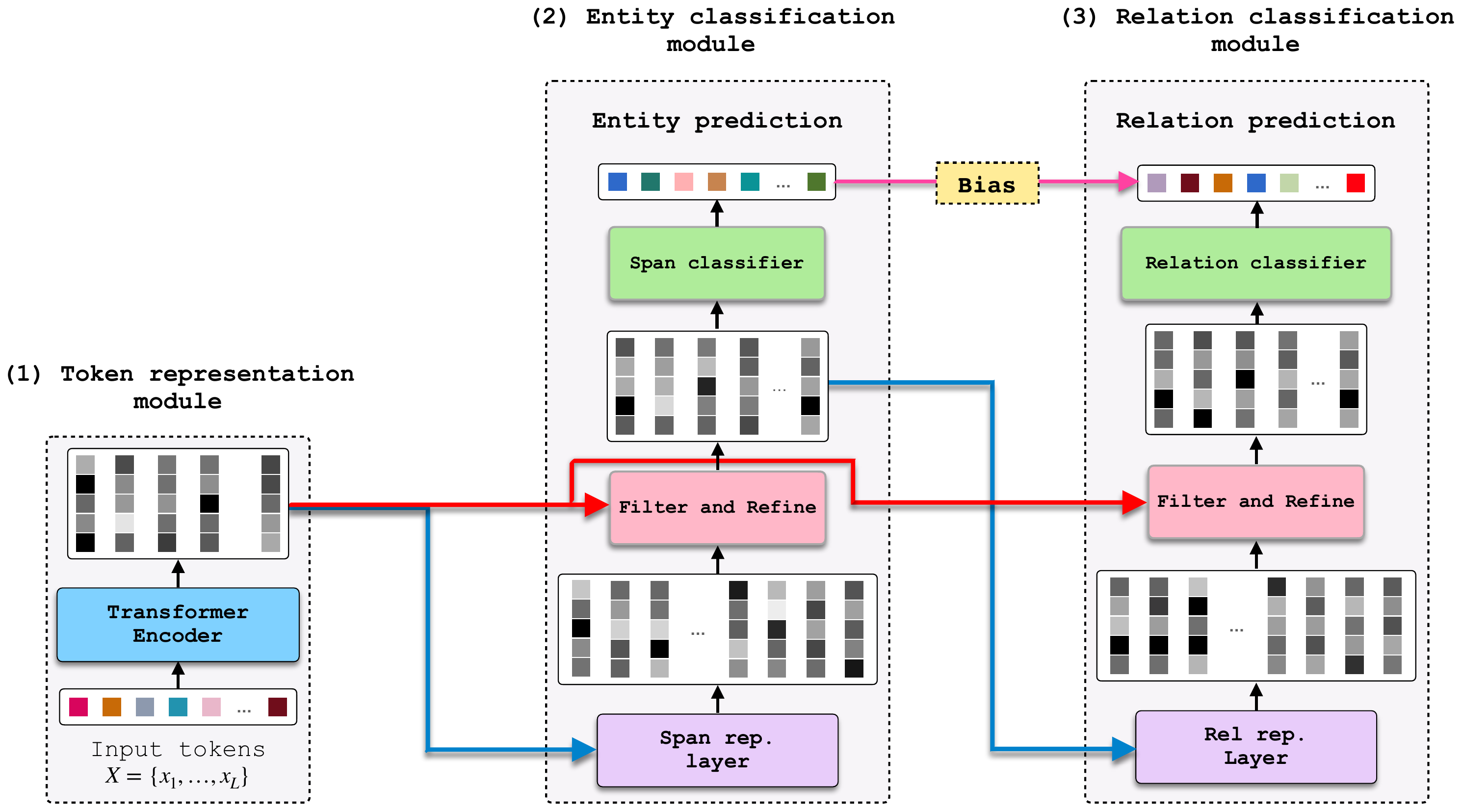}
\caption{
The model consists of three key components: (1) Word Representation, responsible for computing word embeddings for each word in the input sentence. (2) Entity Classification Module, which calculates, prunes, enriches span representations, and classifies them. (3) Relation Classification Module, which similarly calculates, prunes, enriches span representations, and classifies them. The pruning and enrichment of entity and relation representations are performed by a ``Filter and Refine'' layer, as described in Section \ref{sec:refine} and illustrated in Figure \ref{fig:filter}.}
    \label{fig:arch}
\end{figure*}

In this work, we address these limitations by proposing \textit{EnriCo}, a novel framework for joint entity and relation extraction. \textit{EnriCo} aims to provide richer representation and promote coherence in output structures by leveraging attention mechanisms \citep{Vaswani2017AttentionIA} and incorporating task and dataset-specific constraints during decoding. To enhance representation richness, \textit{EnriCo} employs attention mechanisms that allow entities and relations to dynamically attend to relevant parts of the input text (Fig. \ref{fig:attviz}). This allows for richer and more expressive representations by preserving valuable contextual information that may be overlooked by traditional pooling operations. 
In addition, it also incorporates span and relation-level interactions, enabling each candidate entity or relation to update its representation based on the presence and characteristics of other candidate entities or relations in the text. This fosters a more holistic understanding of the relationships between different spans and relations, helping to resolve ambiguities and improve extraction accuracy (Table \ref{abl:attention}). To address the computational complexity arising from a large number of spans and relations, the model integrates a filtering layer to prune candidates, retaining only the most relevant ones, enabling efficient processing without sacrificing accuracy. Previous works have also leveraged rich high-level interactions. For instance, \citet{zaratiana-etal-2022-gnner} employed span-level interaction, yet their application is confined to entity recognition and lacks incorporation of pruning, making it computationally inefficient due to large number of spans. Similarly, \citet{zhu-etal-2023-deep} proposed span-to-token interaction for NER, but our work extends this approach to the relation (span-pair) level.

Finally, to ensure structural coherence of the output, we introduce a series of decoding algorithms to boost model performance by integrating task-specific and dataset-specific constraints. To achieve this, we formulate entity and relation prediction using an Answer Set Programming (ASP) solver, enabling the derivation of exact solutions.
Experiments across benchmark datasets demonstrate the efficacy and performance of our proposed model.
  


\section{Architecture}

In this section, we present the architecture of our proposed model, \textit{EnriCo}, for joint entity and relation extraction. The overall architecture comprises three main components: (1) Word Representation, (2) Entity Classification, and (3) Relation Classification modules. Figure \ref{fig:arch} illustrates the architecture, offering a visual overview of how these modules interact.

\subsection{Token Representation}
\label{sec:word-rep}
The primary purpose of this module is to generate word embeddings from the input sentences. For that, we use a transformer layer that takes an input text sequence $\mathbf{x}$ and outputs token representations $\mathbf{H} \in \mathbb{R}^{L \times D}$, where $D$ is the model dimension. In practice, this component is a pretrained transformer encoder such as BERT \citep{Devlin2019BERTPO}.

\subsection{Entity Module}

In the Entity Module, the objective is to identify and classify spans in the input text as entities. A span refers to a contiguous sequence of words in the text that represents a candidate entity. Each entity is defined by its start and end positions within the input sentence, as well as its associated entity type. For example, the spans ``Alain Farley'' and ``Montreal'' could be classified as entities of type ``Person'' and ``Location'', respectively.

\paragraph{Span representation}

To compute span representations, we first enumerate all possible spans from the input sentence (up to a maximum span length in practice). Then, for each span, we concatenate the embeddings of its start and end words to compute a span representation. More formally, the span representation $\mathbf{S}$ for a span starting at word $i$ and ending at word $j$ is given by:
\begin{equation}
    \mathbf{S}_{ij} = \vw_{ent} ^ T (\mathbf{h}_i^s \oplus \mathbf{h}_j^e)
\label{eq:entrep}
\end{equation}

where $\mathbf{h}_i^s$ and $\mathbf{h}_j^e$ are the embeddings of the start and end words, $\vw_{ent}\in \mathbb{R}^{2D \times D}$ is a learned weight matrix, and $\oplus$ denotes concatenation. In total, we compute $L \times M$ span vectors (we mask invalids), where $L$ represents the sentence length and $M$ represents the maximum span length, thus $\mathbf{S} \in \mathbb{R}^{LM \times D}$. The spans are then passed into a \textit{Filter and Refine} (Sec. \ref{sec:refine}) layer to prune the number of spans to $K$ and update their representation, resulting in $\mathbf{S}_f \in \mathbb{R}^{K \times D}$. The span representation $\mathbf{S}_f$ will serve for both span classification in the next paragraph and the relation representation in Sec. \ref{rel_rep_sec}.

\paragraph{Span classification}

For span classification, we feed the representation of the filtered span $\mathbf{S}_f$ into a feed-forward network to obtain the span classification score:
\begin{equation}
\begin{split}
    \mathbf{Y}^{ent} = \texttt{FFN}(\mathbf{S}_f) \in \mathbb{R}^{K \times |\mathcal{E}|}
\end{split}
\label{eq:entclf}
\end{equation}

where $|\mathcal{E}|$ corresponds to the number of entity types, including the $\texttt{non-entity}$ type.

\subsection{Relation  Module \label{rel_rep_sec}}

In the Relation Module, the goal is to classify pairs of spans in the input text as specific relations. For instance, when presented with two spans ``Alain Farley'' and ``McGill University'', this module has to predict the relation between them, such as ``\textit{Work\_for}'' in this case.

\paragraph{Relation representation}
To compute the representation of a relation between two spans $(i, j)$ and $(k, l)$, we simply concatenate their respective span representations using
\begin{equation}
    \mathbf{R}_{ij|kl} = \vw_{rel}^T (\mathbf{S}_{f_{ij}}^{head} \oplus \mathbf{S}_{f_{kl}}^{tail})
\label{eq:relrep}
\end{equation}

where $\mathbf{S}_{f_{ij}}^{head}$ and $\mathbf{S}_{f_{kl}}^{tail}$ are the span representations for the spans $(i,j)$ and $(k,l)$ respectively and $\vw_{rel}\in \mathbb{R}^{2D \times D}$ is a learned weight matrix. This operation results in $K \times K$ candidate relations, corresponding to all pairs of candidate entities. Similarly to the entities, we process the relation representations through a \textit{Filter and Refine} (Sec. \ref{sec:refine}) layer to reduce their quantity to $K$, thereby updating their representation, which results in $\mathbf{R}_{f} \in \mathbb{R}^{K\times D}$.

\paragraph{Relation classification}
Finally, we compute the relation classification score for each relation representation using a feed-forward network:
\begin{equation}
\begin{split}
\mathbf{Y}^{rel} = \texttt{FFN}(\mathbf{R}_f) \in \mathbb{R}^{K \times |\mathcal{R}|}
\end{split}
\label{eq:relclf}
\end{equation}

where $|\mathcal{R}|$ represents the number of relation types, including the no-relation type.

\subsection{Entity-Relation Biases \label{sec:learned_bias}}

\begin{figure*}[h]
    \centering
\includegraphics[width=0.9\textwidth]{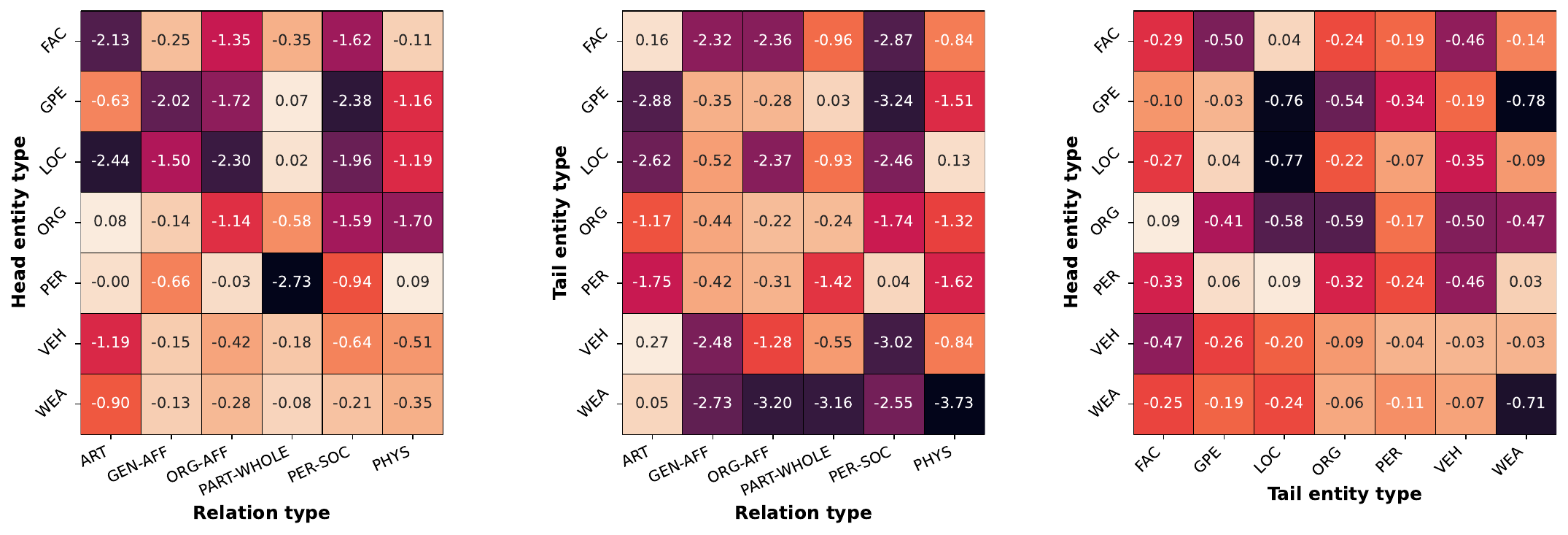}
\caption{\textbf{Biases value (Sec. \ref{sec:learned_bias}) for ACE 05 dataset}. This figure shows the values of learned biases for different associations of entity and relation types. (\textit{left}) $\bm{\phi}(h, r)$, bias scores between head entity type and relation type. (\textit{middle}) $\bm{\phi}(t, r)$, bias scores between tail entity type and relation type. (\textit{right}) $\bm{\phi}(h, t)$, bias scores between head entity type and tail entity type.}
    \label{fig:biasvalue}
\end{figure*}

To facilitate a more nuanced interaction between entity and relation prediction, our model incorporates a bias score for each combination of (head and tail) entity types and relation type (see Fig. \ref{fig:biasvalue} for an illustrative example). In the training phase, these bias scores are learned and seamlessly integrated into the relation score. Specifically, we augment the relation logits (all \(y_r^{rel} \in \mathbf{Y}^{rel}\)) by incorporating information about the predicted head entity type \(h \in \mathcal{E}\) and the tail entity type \(t \in \mathcal{E}\) in the following manner:
\begin{equation}
    y_{rht}^{rel} = y_{r}^{rel} + \textbf{b}(h, t, r)
\end{equation}

where $\textbf{b}(h, t, r) \in \mathbb{R}$, a learned bias score for the triplet $(h, t, r)$ defined as follow:
\begin{equation}
\begin{aligned}
\bm{b}(h, t, r) &= \bm{\phi}(h, t, r)  + \bm{\phi}(h, r)  + \bm{\phi}(t, r)  + \bm{\phi}(h, t)
\end{aligned}
\label{eq:bis}
\end{equation}

In the above, we use the Gumbel-Softmax trick \citep{jang2017categorical} to predict discrete entity types \(h\) and \(t\), enabling gradient-based optimization of the whole process. The term $\bm{\phi}(h, t, r)$ captures the joint affinity score between a specific head, tail, and relation type. 
For instance, if the head entity is \textit{Person} and the tail entity is \textit{Organization}, the relation score would be higher for \textit{works\_for} than \textit{born\_in}. Meanwhile, \( \bm{\phi}(h, r) \) and \( \bm{\phi}(t, r) \) capture the general tendencies for entities (head or tail) of certain types to engage in specific relations. Lastly, \( \bm{\phi}(h, t) \) capture any intrinsic compatibility between an head and tail types. Furthermore, another utility of the bias term is that it allows to incorporate domain constraints by manually assigning large negative values to invalid triples (see Table \ref{const-conll} and \ref{const-ace}).


\begin{figure}[t]
    \centering
    \includegraphics[width=0.9\columnwidth]{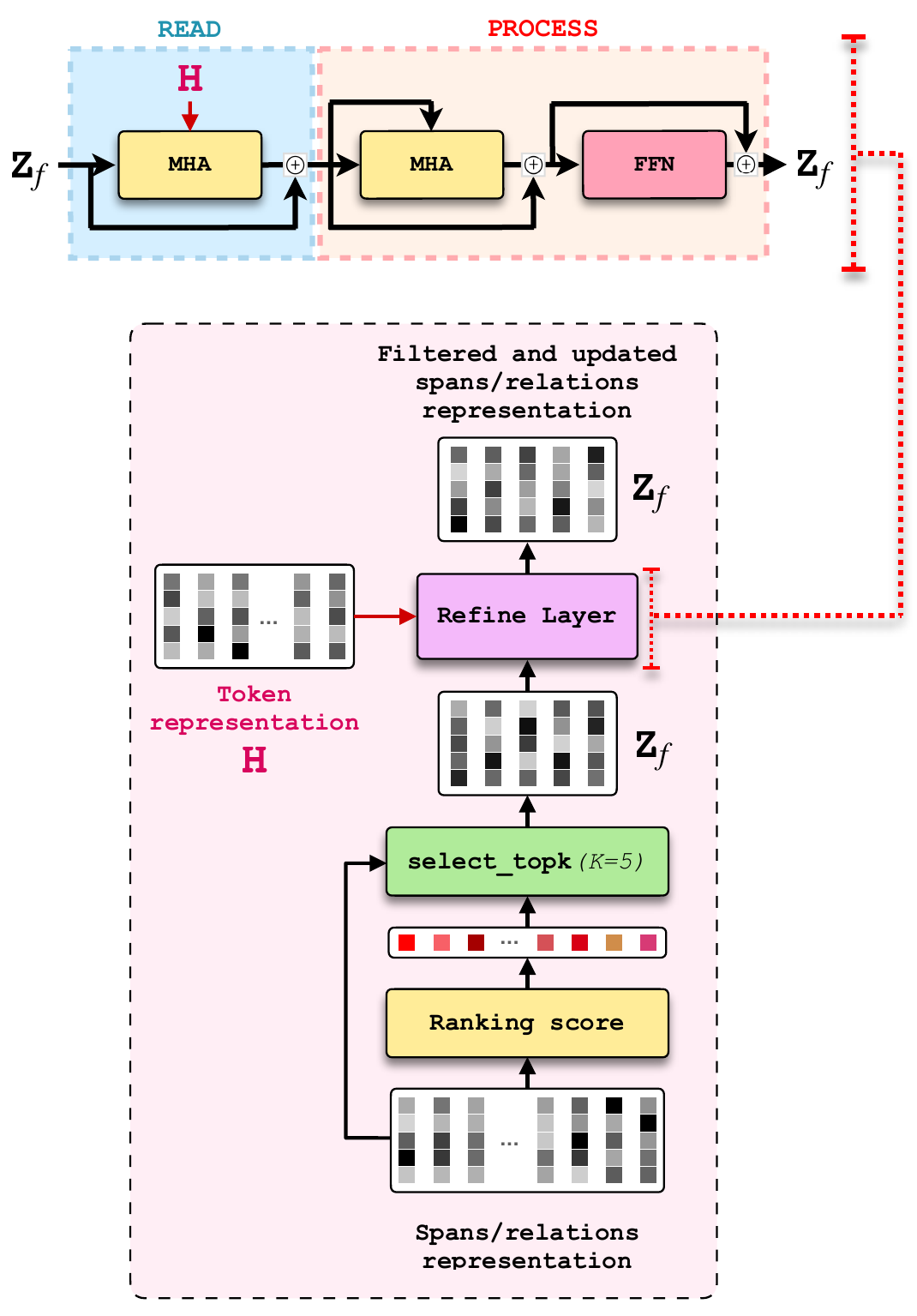}
    \caption{\textbf{Filter and Refine.} This layer processes either span or relation representations. It first computes a ranking score for each span or relation, selecting those with the highest top-k values. The selected spans or relations are then passed through a ``Read \& Process'' layer.}
    \label{fig:filter}
\end{figure}

\subsection{Filter and Refine \label{sec:refine}}

In this section, we detail the \textit{Filter and Refine} layer (see Figure \ref{fig:filter}), a crucial component utilized in both the entity and relation blocks. The purpose of this layer is to prune the candidate elements (entities or relations) and then enhance their representations. Let \(\mathbf{Z} \in \mathbb{R}^{N \times D}\) denote the matrix containing the representations of either entities or relations (i.e., \(\mathbf{S}\) or \(\mathbf{R}\)), where \(N\) represents the number of entities or relations, and \(D\) is the dimension of the model.

\paragraph{Filtering mechanism \label{filtering_mec}}
The filtering first computes ranking scores for each element in \(\mathbf{Z}\) using a FFN:
\begin{equation}
    \mathbf{F} = \texttt{FFN}(\mathbf{Z}) \in \mathbb{R}^{N \times 1}
\end{equation}

\noindent Let \(\texttt{argtopK}(\mathbf{F})\) denote the indices of the top \(K\) elements in the vector \(\mathbf{F}\). Then, the filtered set \(\mathbf{Z}_f \in \mathbb{R}^{K \times D}\) is defined as:
\begin{equation}
\begin{split}
    \mathbf{Z}_f &= \texttt{select\_topk}(\mathbf{Z}, \mathbf{F}) \\
    &= \left[\mathbf{Z}[i] \,|\, i \in \texttt{argtopK}(\mathbf{F})\right]
\end{split}
\end{equation}

This equation defines \(\mathbf{Z}_f\) as the subset of \(\mathbf{Z}\) containing only those elements that are ranked within the top \(K\) according to their scores.

\paragraph{Refine mechanism}

The refine module updates the representations of \(\mZ_f\) using two layers: \texttt{READ} and \texttt{PROCESS}. The \texttt{READ} layer updates each element in \(\mZ_f\) by incorporating information from the original token representation \(\mH\), using multi-head attention:

\begin{equation}
    \mZ_f = \mZ_f + \texttt{MHA}(\mZ_f, \mH)
\end{equation}

\noindent where $\texttt{MHA}(\mathbf{Z}_f, \mathbf{H})$ represents a multi-head attention mechanism with Queries $\mathbf{Z}_f$ and Keys and Values $\mathbf{H}$. This operation proves beneficial as some information may be lost during the hand-crafted representation computation via concatenation (Equations \ref{eq:entrep} and \ref{eq:relrep}). Allowing the entity and relation representations to attend to the original input sequence enables them to dynamically gather crucial information, thereby enhancing overall performance (see ablation study in Table \ref{abl:attention}). Furthermore, this introduces an additional layer of interpretability to our model, as illustrated in Figure \ref{fig:attviz}. Additionally, the \texttt{PROCESS} layer updates the representations by enabling each element ($\in$ \(\mZ_f\)) to aggregate information from others  ($\in$ \(\mZ_f\)):
\begin{equation}
\begin{aligned}
    \mZ_f &= \mZ_f + \texttt{MHA}(\mZ_f, \mZ_f), \\
    \mZ_f &= \mZ_f + \texttt{FFN}(\mZ_f)
\end{aligned}
\end{equation}

\noindent where $\texttt{MHA}(\mathbf{Z}_f, \mathbf{Z}_f)$ is a multi-head self-attention layer, and $\texttt{FFN}(\mathbf{Z}_f)$ is a feed-forward network. While \textit{inter-span} interactions have been explored in previous works \citep{zaratiana-etal-2022-gnner,floquet-etal-2023-attention}, we are the first to employ this mechanism at the relation level.


\subsection{Training}

During training, our model employs multi-task learning by jointly minimizing the filtering and classification losses. We utilize a pairwise ranking loss with margin for the filtering: \cite{10.1145/1553374.1553509}:
\begin{equation}
    \mathcal{L}_f = \sum_{p=1}^{N} \sum_{n=1}^{N} \max(0, \mF_n - \mF_p + \alpha) \cdot \delta(y_p, y_n)
\end{equation}
In this equation, \(\mathbf{F}_p\) and \(\mathbf{F}_n\) represent the filtering scores for positive and negative samples (computed in Sec. \ref{filtering_mec}), respectively. The term \(\alpha\) is the margin, and \(\delta(y_p, y_n)\) is an indicator function that is active when \(y_p = 1\) and \(y_n = 0\). This loss function encourages the model to prioritize positive samples over negative ones. This loss is applied at both the entity and relation levels. For the classification, we minimize the negative log-likelihood of the gold label spans and relations (on $\mathbf{Y}^{ent}$ and $\mathbf{Y}^{rel}$). Finally, the total loss function is a sum of all losses:

\begin{equation}
    \mathcal{L}_{total} = \mathcal{L}_f^{ent} + \mathcal{L}_f^{rel} + \mathcal{L}_{cl}^{ent} + \mathcal{L}_{cl}^{rel}
\end{equation}
Here, $\mathcal{L}_{total}$ is the sum of filtering losses ($\mathcal{L}_f^{ent} + \mathcal{L}_f^{rel}$) and classification losses ($\mathcal{L}_{cl}^{ent} + \mathcal{L}_{cl}^{rel}$) for entities and relations. To maintain simplicity, we do not add weighting terms for individual losses.

\section{Decoding}
\label{sec:decoding}

In this section, we details the different decoding algorithm we employed in this paper. The role of decoding is to produce the final output, which comprises the prediction of entity types (span prediction) and relation types (span pair prediction).

\subsection{Unconstrained Decoding} Our baseline is \textit{unconstrained decoding}, which corresponds to the raw predictions of the model for both entities and relations. The predictions for entities are obtained as follows:

\begin{equation}
E_{\textit{p}} = \left\{(i, j, c) \;\middle|\;
\begin{array}{@{}l@{\;}l}
& c = \argmax_{c'}  \mY_{ijc'}^{ent} \\
& c \neq \texttt{non-entity}
\end{array}
\right\}
\label{eq:filtering_ent}
\end{equation}

where $\mathbf{Y}_{ijc'}^{ent}$ is the score of the spans $(i,j)$ having entity type $c' \in \mathbb{R}^{|\mathcal{E}|}$ (see the computation of span classification score in Equation \ref{eq:entclf}). Furthermore, the prediction of the relations are obtained as follows:

\begin{equation}
R_{\textit{p}} = \left\{(h, t, r) \;\middle|\;
\begin{array}{@{}l@{\;}l}
& r = \argmax_{r'}  \mY_{htr'}^{rel} \\
& r \neq \texttt{no-relation}
\end{array}
\right\}
\label{eq:filtering_rel}
\end{equation}

where $\mY_{htr'}^{rel}$ is the score of the pairs of span $h$ and $t$ having relation type $r' \in \mathbb{R}^{|\mathcal{R}|}$ (the computation of relation classification score is in equation \ref{eq:relclf}). 

\subsection{Constrained Decoding}

\paragraph{Motivations}
The \textit{unconstrained decoding} we describe before, does not consider the task-specific which are crucial for producing well-formed and coherent outputs. For instance, the Joint IE task has the following constraints:

\begin{itemize}
    \item \textbf{Unique Type Assignment:} Each entity and relation must have a unique type assigned to it. (\textit{Trivial})
    \item \textbf{Non-overlapping Entity Spans:} Predicted entity spans must not overlap with each other.
    \item \textbf{Consistency:} A valid relation can only be formed by two valid entities, \textit{i.e.}, a relation cannot be formed by a non-entity span.
\end{itemize}

Moreover, each dataset may have its specific constraints. For instance, in the \textit{CoNLL 04} dataset, if the head entity is \textit{people} and the tail is \textit{Org}, the relation type should be \textit{work\_for} (or non-relation) (see Table \ref{const-conll} for an exhaustive list).

\begin{table}
\centering
\begin{tabular}{llr}
\toprule
\textbf{Head} & \textbf{Tail} & \textbf{Relation} \\
\midrule
\texttt{Peop} & \texttt{Org} & \textit{Work\_For} \\
\texttt{Peop} & \texttt{Loc} & \textit{Live\_in} \\
\texttt{Org} & \texttt{Loc} & \textit{OrgBased\_in} \\
\texttt{Loc} & \texttt{Loc} & \textit{Located\_in} \\
\texttt{Peop} & \texttt{Peop} & \textit{Kill} \\
\bottomrule
\end{tabular}
\caption{CoNLL 04 dataset constraints. The description of the entity and relation types are detailed in the appendix.}
\label{const-conll}
\end{table}

\paragraph{Inference with ASP} In our work, we formulate the decoding problem using \textit{ASP} (Answer Set Programming) \citep{Brewka2011AnswerSP, gebser2014clingo}, a form of declarative programming oriented towards combinatorial search problems. This framework is particularly suitable for our task, as it allows for the integration of various constraints in a straightforward manner. We implement three decoding variants: \underline{Joint}, which jointly optimizes the global score for entities and relations; \underline{Entity First}, which first finds the optimal solution for entities and then for relations conditioned by predicted entities; and \underline{Relation First}, which initially finds the optimal solution for relations and then for entities given the relations. For these decodings, we integrate both task-specific (described above) and dataset constraints (Table \ref{const-conll} and \ref{const-ace}). We provide pseudo-code in the appendix (Figure \ref{asp-code}).

\paragraph{Fast variant} While \textit{ASP} provides strong performance, we find it is slow in practice. To address this, we propose a more scalable solution, which is equivalent to the \underline{Entity First} variant of ASP, described before. Firstly, we predict candidate entities \(E_{\textit{p}}\) using Equation \ref{eq:filtering_ent}. Then, we search for the optimal solution \(\hat{E}_{\textit{p}}\), which is a subset of \(E_{\textit{p}}\) with no overlapping spans and the maximum score:

\begin{equation}
\hat{E}_{\textit{p}} = \argmax_{E \in \Psi(E_{\textit{p}})} \sum_{(i,j,c)\in E} \mY_{ijc}^{ent}
\label{eq:decode}
\end{equation}

where $\Psi(E_{\textit{p}})$ contains all possible solution. The solution to this problem is provided by \citet{zaratiana-etal-2022-named}, who transform the problem into a weighted graph search to derive \textit{exact solution}. Then, once the entities are determined, the goal is to predict the types of each candidate relation based on these predicted entities. A key assumption is that the type of one relation is independent of others, provided the entities are known (i.e there is no \textit{inter-relation} constraints). Therefore, we can predict each relation types (for all $y^{rel} \in \mY^{rel}$) independently as follow:
\begin{equation}
    r = \argmax_{r \in \mathcal{R}} y_{r}^{rel} + \textbf{b}(h, t, r)
\end{equation}

where $h$ and $t$ are respectively the type of head and tail entities. The bias term $\textbf{b}(h, t, r)$ add entity prediction information in the relation facilitate the integration of constraints into the prediction. It does so by assigning a negative infinity value to any invalid entity-relation type associations, as dictated by the specific dataset constraint (Table \ref{const-conll} and \ref{const-ace}). As shown in the table \ref{tab:decoding_speed}, this algorithm is significantly faster than ASP-based approached, while allowing the adherence to constraints.

\begin{table}[]
\centering
\label{table:ace_constraints}
\resizebox{\columnwidth}{!}{
\begin{tabular}{lll}
\toprule
\textbf{Head} & \textbf{Tail} & \textbf{Relations} \\
\midrule
\texttt{PER} & \texttt{FAC} & \textit{ART, PHYS} \\
\texttt{PER} & \texttt{LOC} & \textit{PHYS, GEN-AFF} \\
\texttt{PER} & \texttt{GPE} & \textit{PHYS, ORG-AFF, GEN-AFF} \\
\texttt{PER} & \texttt{PER} & \textit{PER-SOC, GEN-AFF} \\
\texttt{PER} & \texttt{ORG} & \textit{ORG-AFF, GEN-AFF} \\
\texttt{PER} & \texttt{WEA} & \textit{ART} \\
\texttt{PER} & \texttt{VEH} & \textit{ART} \\
\texttt{FAC} & \texttt{FAC} & \textit{PART-WHOLE, PHYS} \\
\texttt{FAC} & \texttt{GPE} & \textit{PART-WHOLE, PHYS} \\
\texttt{FAC} & \texttt{LOC} & \textit{PART-WHOLE, PHYS} \\
\texttt{GPE} & \texttt{FAC} & \textit{PART-WHOLE, PHYS, ART} \\
\texttt{GPE} & \texttt{GPE} & \textit{PART-WHOLE, PHYS, ORG-AFF} \\
\texttt{GPE} & \texttt{LOC} & \textit{PART-WHOLE, PHYS} \\
\texttt{GPE} & \texttt{ORG} & \textit{ORG-AFF} \\
\texttt{GPE} & \texttt{WEA} & \textit{ART} \\
\texttt{GPE} & \texttt{VEH} & \textit{ART} \\
\texttt{LOC} & \texttt{FAC} & \textit{PART-WHOLE, PHYS} \\
\texttt{LOC} & \texttt{GPE} & \textit{PART-WHOLE, PHYS} \\
\texttt{LOC} & \texttt{LOC} & \textit{PART-WHOLE, PHYS} \\
\texttt{ORG} & \texttt{ORG} & \textit{PART-WHOLE, ORG-AFF} \\
\texttt{ORG} & \texttt{GPE} & \textit{PART-WHOLE, ORG-AFF, GEN-AFF} \\
\texttt{ORG} & \texttt{WEA} & \textit{ART} \\
\texttt{ORG} & \texttt{VEH} & \textit{ART} \\
\texttt{ORG} & \texttt{FAC} & \textit{ART} \\
\texttt{ORG} & \texttt{LOC} & \textit{GEN-AFF} \\
\texttt{VEH} & \texttt{VEH} & \textit{PART-WHOLE} \\
\texttt{WEA} & \texttt{WEA} & \textit{PART-WHOLE} \\
\bottomrule
\end{tabular}
}
\caption{ACE 05 dataset constraints. The description of the entity and relation types are detailed in the appendix.}
\label{const-ace}
\end{table}

\begin{table}[h]
    \centering
    \begin{tabular}{@{}lcc@{}}
        \toprule
                            & \textbf{ASP-based} & \textbf{Fast variant} \\ 
        \midrule
        Joint               & 6.7              & -                    \\
        Relation first      & 7.4              & -                    \\
        Entity first        & 5.5              & 21.7                 \\
        \bottomrule
    \end{tabular}
    \caption{Decoding speed in sentence per second. All decoding can be implemented using ASP solver. Entity first variant can be implemented without ASP resulting in faster decoding.}
    \label{tab:decoding_speed}
\end{table}

\section{Experimental Setup}
\subsection{Datasets}
We evaluated our model on three datasets for joint entity-relation extraction, namely SciERC \citep{luan2018multitask}, CoNLL04 \citep{carreras-marquez-2004-introduction}, and ACE 05 \citep{ace05}. We provide details and statistics about the datasets in the Table \ref{tab:dataset_statistics} and the description of entity and relation types in Table \ref{tab:type_desc}.

\paragraph{\textbf{ACE 05}} is collected from a variety of domains, such as newswire, online forums and broadcast news. It provides a diverse set of entity types such as Persons (PER), Locations (LOC), Geopolitical Entities (GPE), and Organizations (ORG), along with intricate relation types that include Artifact relationships (ART), General affiliations (GEN-AFF), and Personal social relationships  (PER-SOC). This dataset is particularly notable for its complexity and wide coverage of entity and relation types, making it a robust benchmark for evaluating the performance of Joint IE models.

 \paragraph{\textbf{CoNLL04}} is a popular benchmark dataset for entity-relation extraction in English. It focuses on general entities such as People, Organizations, and Locations. The dataset primarily includes simple and generic relations like \textit{Work\_For} and \textit{Live\_in}.

\paragraph{\noindent\textbf{SciERC}} dataset is specifically designed for the AI domain. It includes entity and relation annotations from a collection of documents from 500 AI paper abstracts. It contains entity types such as Task, Method, Metric and relation types such as \textit{Use-for}, \textit{Part-of} and \textit{Compare}. SciERC is particularly suited for constructing knowledge graphs in the AI domain.

\subsection{Dataset Constraints}
In this section, we discuss the dataset constraints used in our work. For the CoNLL 04 dataset, the constraints are based on the seminal work of Roth and Yih \cite{roth-yih-2004-linear}, which we report in Table \ref{const-conll}. The constraints for this dataset are relatively simple, allowing only five triplet combinations, for instance, (Peop, Org, Work\_For). For the ACE 05 dataset, no constraints were publicly available. Thus, we decided to design the constraints manually by examining the annotation guidelines provided by the Linguistic Data Consortium dataset's annotation guidelines \footnote{https://www.ldc.upenn.edu/collaborations/past-projects/ace/annotation-tasks-and-specifications}, resulting in the set of constraints reported in Table \ref{const-ace}. As shown in the table, the task for the ACE 05 dataset is highly complex, with more than 40 possible triples compared to CoNLL 04, which only has 5. Finally, for SciERC, we do not include dataset-specific constraints as the annotation guideline is not detailed enough to permit that, and the presence of ill-defined entities such as \textit{Generic} and \textit{Other-ScientificTerm} makes it difficult (see Table \ref{tab:type_desc}).

\begin{table}[]
\centering
\resizebox{\columnwidth}{!}{
\begin{tabular}{@{}lcccccc@{}}
\toprule
Dataset & \(|\mathcal{E}|\) & \(|\mathcal{R}|\) & \# Train & \# Dev & \# Test \\
\midrule
ACE05 & 7 & 6 & 10,051 & 2,424 & 2,050 \\
CoNLL 04 & 4 & 5 & 922 & 231 & 288 \\
SciERC & 6 & 7 & 1,861 & 275 & 551 \\
\bottomrule
\end{tabular}
}
\caption{The statistics of the datasets. We use ACE04, ACE05, SciERC, and CoNLL 04 for evaluating end-to-end relation extraction.}
\label{tab:dataset_statistics}
\end{table}

\begin{table}[]
\centering
\label{tab:hyperparameters}
\begin{tabular}{@{}lccc@{}}
\toprule
 & ACE 05 & CoNLL 04 & SciERC \\ 
\midrule
Backbone      & ALB    &  ALB  & SciB  \\
Optimizer      &  & AdamW  &    \\
lr backbone     & 1e-5  &    3e-5        &   3e-5     \\
lr others         &   5e-5     &   5e-5         & 5e-5       \\
Weight Decay   &   1e-2     &  1e-4          &  1e-4      \\
Dropout    &  0.1      &   0.1         &   0.1     \\
Hidden Size    &  768      &  768          &    768    \\
Train steps &  120k     &  50k          &   20k \\ Warmup         &  5k      &   5k         &  2k      \\Batch size &  8     &  8          &   8    \\ Span length &  12     &  12          &   12   \\ 
\bottomrule
\end{tabular}
\caption{\textbf{Hyperparameters}.}
\label{tab:hyper}
\end{table}

\begin{table*}[]
    \centering
    \label{tab:comparison}
    \begin{tabular}{l|l|ccc|ccc|ccccccccc}
        \toprule
        & & \multicolumn{3}{c|}{\textbf{ACE 05}} & \multicolumn{3}{c|}{\textbf{CoNLL 04}} & \multicolumn{3}{c}{\textbf{SciERC}} \\
        Model & Backbone & ENT & REL & REL+ & ENT & REL & REL+ & ENT & REL & REL+ \\
        \midrule
        DYGIE++ & BB \& SciB & 88.6 & 63.4 & -- & -- & -- & -- & 67.5 & \underline{48.4} & -- \\
        Tab-Seq  & ALB & 89.5 & -- & 64.3 & 90.1 & 73.8 & 73.6 & -- & -- & -- \\
        PURE & ALB \& SciB & 89.7 & 69.0 & 65.6 & -- & -- & -- & 66.6 & 48.2 & 35.6 \\
        PFN & ALB \& SciB & 89.0 & -- & 66.8 & -- & -- & -- & 66.8 & -- & 38.4 \\
        UniRE & ALB \& SciB & 89.9 & -- & 66.0 & -- & -- & -- & \underline{68.4} & -- & 36.9 \\
        TablERT & ALB & 87.8 & 65.0 & 61.8 & \textbf{90.5} & 73.2 & 72.2 & -- & -- & -- \\
        UTC-IE & ALB \& SciB & 89.9 & -- & \textbf{67.8} & -- & -- & -- & 69.0 & -- & \underline{38.8} \\
        UIE  & T5          & -- & -- & 66.6 & -- & 75.0 & -- & -- & -- & 36.5 \\
        ChatGPT &     (\textit{Few-shot})      & -- & 9.04 & -- & -- & \underline{76.5} & -- & -- & 17.92 & -- \\
        \midrule
        \multirow{2}{*}{\textit{EnriCo}} & ALB \& SciB & \textbf{90.1} & \textbf{69.1} & \underline{67.6} & 89.8 & \textbf{76.6} & \textbf{76.6} &\textbf{69.3} & \textbf{50.5} & \textbf{40.2} \\
                                             & BL & 88.8 & 67.2 & 64.7 & 89.9 & 73.5 & 73.5 & -- & -- & -- \\
        \bottomrule
    \end{tabular}
    
    \caption{\textbf{Main results}. \textit{Entity} refers to the F1 score for entity recognition, \textit{REL} for relaxed relation extraction, and \textit{REL+} for strict relation extraction. The \textit{Backbone} column indicates the underlying architecture for each model (ALB for \texttt{albert-xxlarge-v1} \citep{Lan2019ALBERTAL}, BL for \texttt{bert-large-cased} \citep{Devlin2019BERTPO}, and SciB for \texttt{scibert-base-uncased} \citep{Beltagy2019SciBERTAP}).}
\end{table*}

\subsection{Evaluation Metrics}

For the named entity recognition (NER) task, we use span-level evaluation, demanding precise entity boundary and type predictions. In evaluating relations, we employ two metrics: (1) Boundary Evaluation (REL), which requires correct prediction of entity boundaries and relation types, and (2) Strict Evaluation (REL+), which also necessitates correct entity type prediction. We report the micro-averaged F1 score following previous works.

\subsection{Hyperparameters}
In this study, we implemented our model using BERT \citep{Devlin2019BERTPO} or ALBERT \citep{Lan2019ALBERTAL} for the CoNLL 04 and ACE 05 datasets. For the SciERC dataset, we opted for SciBERT \citep{Beltagy2019SciBERTAP}, aligning with previous works. We detail the hyperparameters in Table \ref{tab:hyper}. Our model was implemented using PyTorch and trained on a server equipped with A100 GPUs.

\subsection{Baselines} We primarily compare our model, \textit{EnriCo}, with comparable approaches from the literature in terms of model size. \textbf{DyGIE++} \citep{wadden-etal-2019-entity} is a model that uses a pretrained transformer to compute contextualized representations and employs graph propagation to update the representations of spans for prediction. \textbf{PURE} \citep{zhong-chen-2021-frustratingly} is a pipeline model for the information extraction task that learns distinct contextual representations for entities and relations. \textbf{PFN} \citep{yan-etal-2021-partition} introduces methods that model two-way interactions between the task by partitioning and filtering features. \textbf{UniRE} \citep{wang-etal-2021-unire} proposes a joint entity and relation extraction model that uses a unified label space for entity and relation classification. \textbf{Tab-Seq} \citep{wang-lu-2020-two} tackles the task of joint information extraction by treating it as a table-filling problem. Similarly, in \textbf{TablERT} \citep{ma-etal-2022-joint}, entities and relations are treated as tables, and the model utilizes two-dimensional CNNs to effectively capture and model local dependencies within these table-like structures. Finally, \textbf{UTC-IE} \citep{yan-etal-2023-utc} treats the task as token-pair classification. It incorporates Plusformer to facilitate axis-aware interactions through plus-shaped self-attention and local interactions via Convolutional Neural Networks over token pairs. We also included evaluations of generative approaches for information extraction, comprising \textbf{UIE} \citep{lu-etal-2022-unified}, which fine-tunes a T5 model for information extraction, and \textbf{ChatGPT} \citep{wadhwa-etal-2023-revisiting} prompted using few-shot demonstrations.

\section{Results and Analysis}
\subsection{Main Results}
The main results of our experiments are reported in Table 2. On ACE 05, our model obtains the highest results in entity evaluation and is second in relation prediction, slightly under-performing \textit{UTC-IE}. On CoNLL 04, our model surpasses the best \textit{non-generative baseline} by a large margin. Specifically, it obtains 76.6 on relation evaluation, achieving a +3 F1 improvement compared to \textit{Tab-Seq}. Similarly, on SciERC, it also obtains strong results for both entities and relations, outperforming \textit{UTC-IE} by 0.3 and 1.4 on entity and relation F1, respectively. Furthermore, our model also show competitive performance against generative models, \textit{UIE} and \textit{ChatGPT}. On CoNLL 04, \textit{ChatGPT} performs quite well due to the simplicity of relations in this dataset. However, on more complex datasets (SciERC and ACE 05), its performance is far behind, showing the benefits of fine-tuning task-specific models for the task. Overall, our model showcases strong performance across all datasets, demonstrating the utility of our proposed framework.

\begin{table}[]
    \centering
    \begin{tabular}{llcccc}
        \toprule
        Dataset & Setting & ENT & REL & REL+ \\
        \midrule
        \textbf{ACE 05} & \texttt{Unconstr.}   & 60.2 & 68.6 & 67.1 \\
        \cmidrule(lr){2-5}
               & \texttt{Joint}  & 90.1 & 68.9 & 67.5 \\
               & \texttt{Ent First}  & \textbf{90.2} & \textbf{69.1} & \textbf{67.6} 
               \\
               & \texttt{Rel First} & 90.0 & 68.7 & 66.9 \\
        \midrule
        \textbf{CoNLL 04} & \texttt{Unconstr.}   & 60.6 & 76.3 & 76.1 \\
        \cmidrule(lr){2-5}
               & \texttt{Joint}  & \textbf{89.8} & 76.6 & 76.6 \\
               & \texttt{Ent First}  & \textbf{89.8} & \textbf{76.7} & \textbf{76.7} \\
               & \texttt{Rel First} & 89.7 & 76.5 & 76.5 \\ 
        \midrule
        \textbf{SciERC} & \texttt{Unconstr.}   & \textbf{69.4} & \textbf{51.4} & 39.7 \\ \cmidrule(lr){2-5}
               & \texttt{Joint}  & 69.3 & 50.8 & 40.5 \\
               & \texttt{Ent First}  & 69.3 & 50.5 & 40.2 \\
               & \texttt{Rel First} & \textbf{69.4} & 51.1 & \textbf{40.7} \\ 
        \bottomrule
    \end{tabular}
    \caption{\textbf{Performance Comparison of Decoding Algorithms.} We compare unconstrained and constrained approaches.}
    \label{tab:decoding_res}
\end{table}

\subsection{Decoding Algorithms}
In Table \ref{tab:decoding_res}, we report the performance of our model using different decoding algorithms described in Section \ref{sec:decoding}. We observe that, as expected, \textit{unconstrained decoding} is the least competitive, except on SciERC where we did not apply a domain constraint. In particular, \textit{unconstrained decoding} performance on entity recognition can be very poor, especially for ACE 05 and CoNLL 04, where it falls behind the constrained method by almost 30 points in terms of F1, mainly due to span boundary and span overlap errors. For relation extraction, constrained decoding can improve by up to 0.5, 0.6, and 1.0 points in terms of the F1 score on ACE 05, CoNLL 04, and SciERC, respectively. These results demonstrate that structural and domain constraints are important not only for improving coherence but also for performance. Furthermore, we notice that the performance difference between different constrained decoding methods (\textit{Joint}, \textit{Entity First}, and \textit{Relation First}) is minimal across datasets. However, \textit{Entity First} is the most beneficial one as it can be implemented efficiently without the need for using an \textit{ASP} solver, making it up to 3x to 4x faster than other alternatives (Table \ref{tab:decoding_speed}).


\begin{table}
    \centering
    \begin{tabular}{llcccc}
        \toprule
        Dataset & Setting & ENT & REL & REL+ \\
        \midrule
        \textbf{ACE 05} & \texttt{Full.}   & \textbf{90.1} & \textbf{69.1} & \textbf{67.6} \\
               & \texttt{No ent.}  & 89.7 & 67.6 & 65.9 \\
               & \texttt{No rel.}  & \textbf{90.2} & \underline{68.1} & \underline{66.5} \\
               & \texttt{No both} & 89.7 & 67.6 & 65.7 \\
        \midrule
        \textbf{CoNLL 04} & \texttt{Full.}   & \textbf{89.8} & \textbf{76.6} & \textbf{76.6} \\
                   & \texttt{No ent.}  & 89.7 & 76.1 & 76.1 \\
                   & \texttt{No rel.}  & \textbf{89.8} & \underline{76.4} & \underline{76.4} \\
                   & \texttt{No both} & 89.5 & 75.7 & 75.7 \\
        \midrule
        \textbf{SciERC} & \texttt{Full.}   & \underline{69.3} & \textbf{50.5} & \textbf{40.2} \\
               & \texttt{No ent.}  & \textbf{69.8} & \underline{50.4} & \underline{39.8} \\
               & \texttt{No rel.}  & 69.0 & 50.3 & 39.6 \\
               & \texttt{No both} & 68.0 & 50.0 & 38.8 \\
        \bottomrule
    \end{tabular}
    \caption{\textbf{Ablation experiment}. With and without refine layer at the entity/relation level.}
\label{abl:attention}
\end{table}

\subsection{Refine Layer Ablation} We perform an ablation analysis in Table \ref{abl:attention} to assess the effectiveness of the refine layer, specifically examining the contributions of the entity-level and relation-level refine layers described in Section \ref{sec:refine}. To ensure a fair comparison, we maintain a similar number of parameters for all compared variants. In general, our model with the full configuration—incorporating both entity and relation level interactions—achieves the most competitive scores across the datasets. However, removing either the entity or relation level interaction does not significantly impact performance, whereas removing both leads to a more substantial drop in performance.

\begin{figure*}[h]
    \centering
    \includegraphics[width=\textwidth]{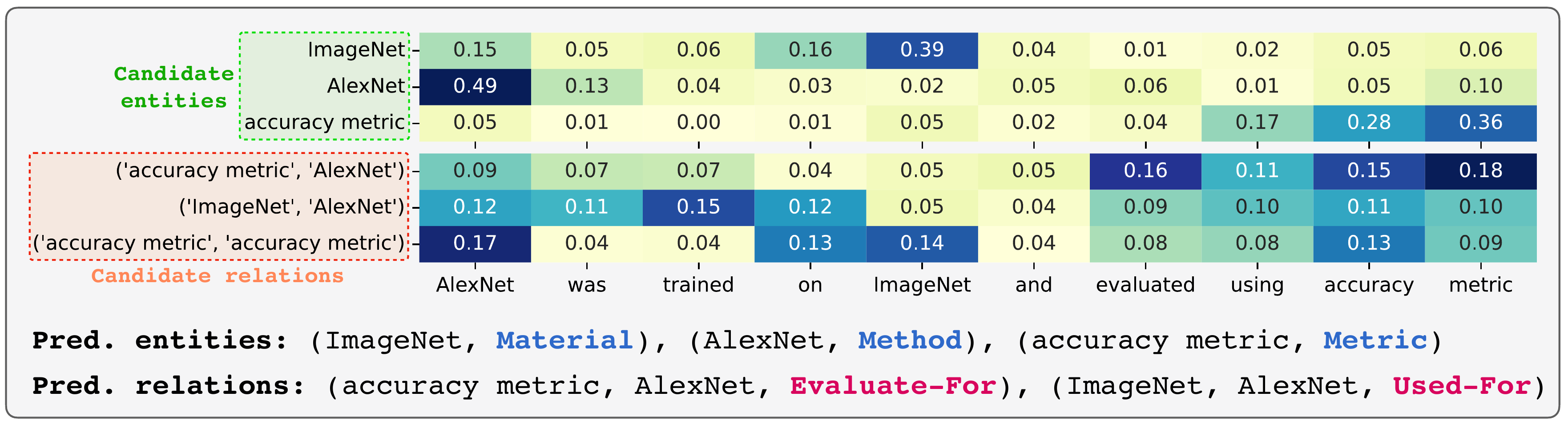} 
    \caption{\textbf{Attention visualization}. This illustrates the attention scores of candidate entities and candidate relations within the input sequence, averaged across attention heads.} %
    \label{fig:attviz}
\end{figure*}

\subsection{Attention Visualization} In Figure \ref{fig:attviz}, we present the attention visualization of the \texttt{READ} module for entities and relations, highlighting their interaction with the input sequence. This visualization depicts the attention scores averaged across all attention heads. The examples illustrated demonstrate that each span generally attends most to its corresponding position in the input text. However, intriguingly, we also observe attention to certain clue words such as ``on'' and ``using'', which may contribute to type prediction. For relations, attention is directed to both head and tail spans constituting the relation. However, contextual information beyond the spans is attended to; for example, the word ``evaluated'' receives significant attention from the (``accuracy metric'', ``AlexNet'') relation, indicating the Evaluate-For relation between the two spans. Similarly, in the same line, the word ``trained'' is highly attended to by the (``ImageNet'', ``AlexNet'') pair.

\section{Related Works}
\paragraph{Joint IE} The field of information extraction (IE) has evolved from traditional pipeline models, which sequentially handle entity recognition \citep{Chiu2015NamedER, Lample2016NeuralAF} and relation extraction \citep{Zelenko2002KernelMF, Bach2007ARO, Lin2016NeuralRE, Wu2017AdversarialTF}, to end-to-end models. These approaches aim to mitigate error propagation \citep{10.1007/10704656_11, Nadeau2007ASO} by jointly optimizing entity and relation extraction \citep{roth-yih-2004-linear,Fu2019GraphRelMT, Sun_Zhang_Mensah_Mao_Liu_2021, ye-etal-2022-packed}, enhancing the interaction between the two task and overall performance. Proposed approaches include table-filling methods \citep{wang-lu-2020-two, ma-etal-2022-joint}, span pair classification \citep{Eberts2019SpanbasedJE, wadden-etal-2019-entity}, set prediction \citep{Sui2020JointEA}, augmented sequence tagging \citep{ji-etal-2020-span} and the use of unified labels for the task \citep{wang-etal-2021-unire, yan-etal-2023-utc}. In addition, recently, the usage of generative models \citep{Achiam2023GPT4TR} has become popular for this task, where input texts are encoded and decoded into augmented language \citep{paolini2021structured}. Some of these approaches conduct fine-tuning on labeled datasets \citep{lu-etal-2022-unified,fei2022lasuie,Zaratiana_Tomeh_Holat_Charnois_2024}, and others prompt large language models such as ChatGPT \citep{wadhwa-etal-2023-revisiting}.

\paragraph{Higher-order attention} Recent works have proposed higher-order interactions for structured prediction models. For instance, \citet{floquet-etal-2023-attention} employed span-level attention for parsing, utilizing linear transformers to circumvent quadratic complexity of dot-product attention. The work of \citet{zaratiana-etal-2022-gnner} employed span-level Graph Attention Networks \citep{Velickovic2017GraphAN} to enhance span representations for Named Entity Recognition (NER), using overlap information as edges. However, their approach is slow and takes huge memory due to the substantial size of the overlap graph, characterized by numerous nodes and edges. In our work, we address this challenge by implementing a filtering mechanism to alleviate computational inefficiencies. Similarly, \citet{ji-etal-2023-improving} leveraged span-level attention by restricting the number of attended spans for each span using predefined heuristic. In contrast, our proposed method dynamically selects them. Additionally, \citet{zhu-etal-2023-deep} utilized span-to-token attention for Named Entity Recognition (NER). Our model extends their approach by incorporating both span- and relation-level interaction.

\section{Conclusion}
In summary, this paper introduces \textit{EnriCo}, a novel model crafted for joint entity and relation extraction tasks. By integrating span-level and relation-level attention mechanisms, our model fosters richer representations of spans and their interactions. The incorporation of a filtering mechanism efficiently manages computational complexity, while the integration of learned biases and constraint-based decoding further enhances the precision of model predictions. Experimental evaluations across benchmark datasets demonstrate the efficacy and performance of our proposed model.

\section*{Acknowledgments}  This work was granted access to the HPC resources of IDRIS under the allocation 2023-AD011014472 and AD011013682R1 made by GENCI. This work is partially supported by a public grantoverseen by the French National Research Agency
(ANR) as part of the program Investissements
d’Avenir (ANR-10-LABX-0083).

\bibliography{custom,ant}

\appendix

\begin{table*}[]
\centering
\label{table:combined_descriptions}
{\fontsize{12pt}{14pt}\selectfont
\begin{tabular}{@{}lll@{}}
\toprule
Dataset & Type & Description \\ \midrule
\textbf{ACE 05} & \multicolumn{2}{l}{\textbf{Entity Types}} \\
& PER & Individual people or groups of people \\
& FAC & Buildings, airports, highways, bridges, etc. \\
& LOC & Geographical areas and landmasses \\
& GPE & Countries, cities, states, etc. \\
& ORG & Companies, governments, institutions, etc. \\
& WEA & Tools or instruments designed to inflict harm \\
& VEH & Means of transportation \\
& \multicolumn{2}{l}{\textbf{Relation Types}} \\
& ART & Artifact relationship, indicating ownership or use \\
& PHYS & Physical proximity or location relationships \\
& GEN-AFF & General affiliations or relationships \\
& ORG-AFF & Affiliation with an organization \\
& PER-SOC & Personal relationships between people \\
& PART-WHOLE & One entity is a part of another entity \\
\addlinespace
\midrule
\textbf{CoNLL 2004} & \multicolumn{2}{l}{\textbf{Entity Types}} \\
& Peop & Individuals or groups of people \\
& Org & Companies, institutions, governments, etc. \\
& Loc & Geographical entities like cities, countries, etc. \\
& \multicolumn{2}{l}{\textbf{Relation Types}} \\
& Work\_For & A person works for an organization \\
& Live\_in & A person lives in a location \\
& OrgBased\_in & An organization is based in a location \\
& Located\_in & One location is situated within another \\
& Kill & One person is responsible for the death of another \\
\addlinespace
\midrule
\textbf{SciERC} & \multicolumn{2}{l}{\textbf{Entity Types}} \\
& Task & Specific task or problem in a scientific context \\
& Method & Approach or technique used in research \\
& Metric & Standard of measurement in experiments \\
& Material & Tools, datasets, or materials used in research \\
& Other-ScientificTerm & Other relevant scientific terms \\
& Generic & General entities relevant in context \\
& \multicolumn{2}{l}{\textbf{Relation Types}} \\
& Used-for & A method/material used for a task \\
& Feature-of & An entity is a feature of another \\
& Part-of & One entity is a component of another \\
& Evaluate-for & Evaluating a method/material/task \\
& Compare & Comparison between two or more entities \\
& Conjunction & Entities related in a conjunctive manner \\
\bottomrule
\end{tabular}
}
\caption{Combined Descriptions of Entity and Relation Types in ACE 05, CoNLL 2004, and SciERC datasets.}
\label{tab:type_desc}
\end{table*}


\begin{figure*}[]
    \centering
\includegraphics[width=1.\textwidth]{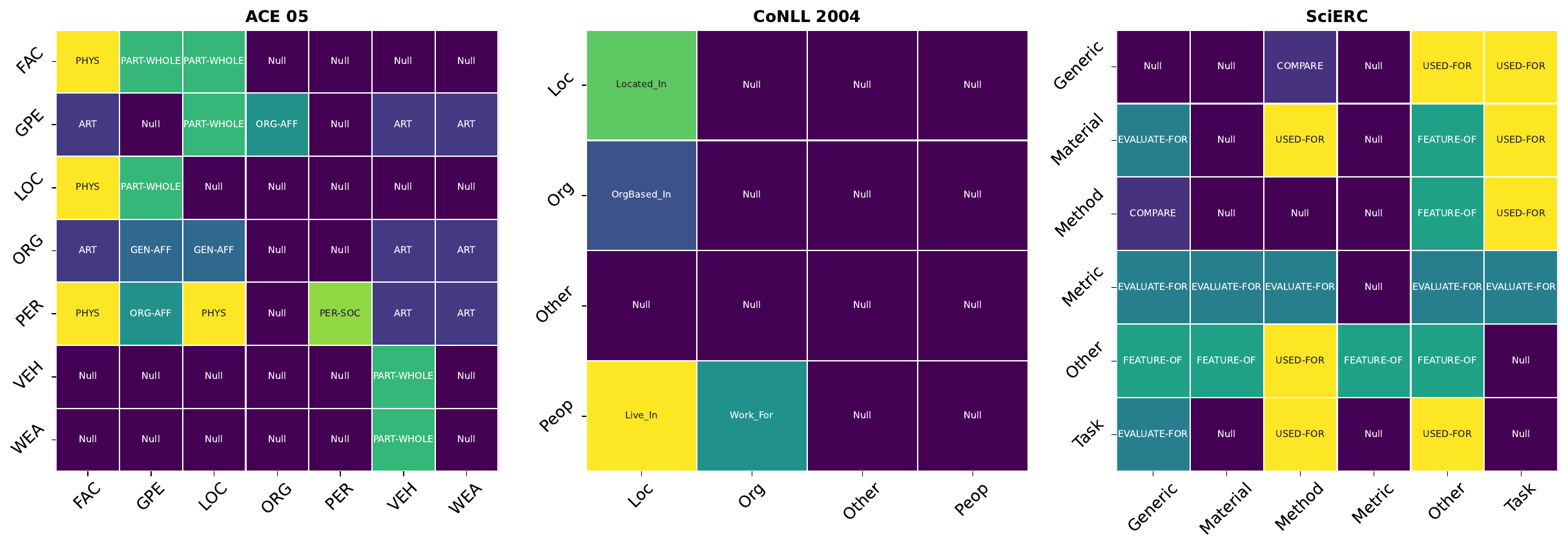}
\caption{This figure illustrate the highest scoring relation type for each pairs of entity types.}
    \label{fig:learnedbias}
\end{figure*}

\begin{figure*}[h]
\vspace{1.5em}
    \centering
\includegraphics[width=1.\textwidth]{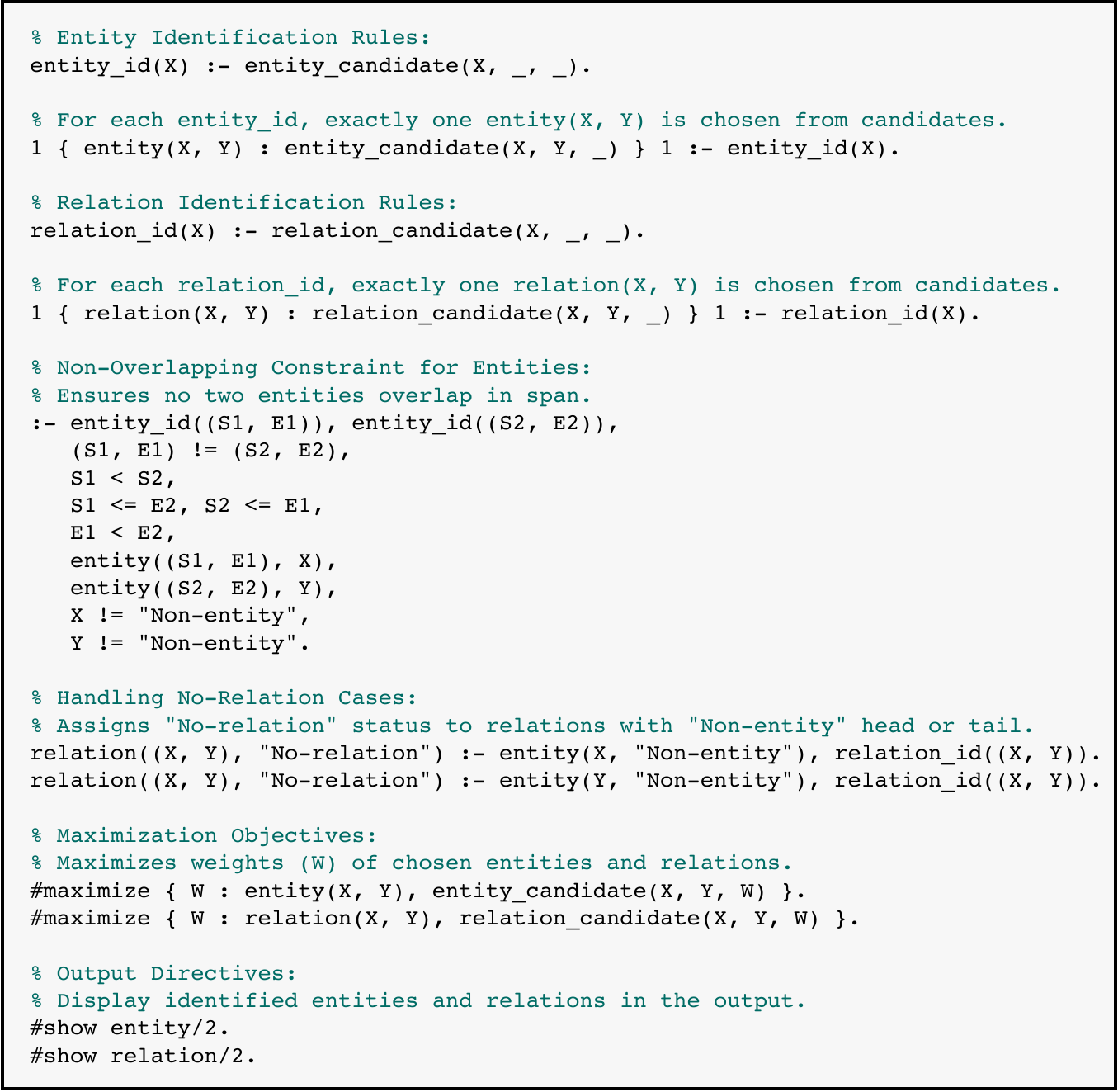}
\caption{ASP pseudo code for globally constrained decoding for joint entity and relation extraction.}
\label{asp-code}
\end{figure*}

\end{document}